\documentclass{article}

\usepackage{arxiv}

\usepackage[T1]{fontenc}
\usepackage[utf8]{inputenc}
\usepackage{amsmath,amssymb}
\usepackage[english]{babel}
\usepackage{hyperref}  
\usepackage{url}     
\usepackage{booktabs}      
\usepackage{amsfonts}  
\usepackage{nicefrac}  
\usepackage{microtype}   
\usepackage{graphicx}
\usepackage{caption}
\usepackage{natbib}
\usepackage{doi}
\usepackage{authblk}
\usepackage{multirow}
\usepackage{array}
\usepackage[table]{xcolor}

\newcolumntype{C}[1]{>{\centering\let\newline\\\arraybackslash\hspace{0pt}}m{#1}}

\setcitestyle{numbers,open={[},close={]}}

\title{Homeostazis and Sparsity in Transformer}

\date{}

\renewcommand{\shorttitle}{Homeostasis and Sparsity in Transformer}

\hypersetup{
pdftitle={HOMEOSTASIS AND SPARSITY IN TRANSFORMER},
pdfsubject={cs.AI},
pdfauthor={Leonid Kotyuzanskiy, Artem Klimov},
pdfkeywords={transformer, sparse distributed representations, homeostasis, natural language processing, machine translation},
}

\begin{document}
\author[1, 2]{\small Leonid Kotyuzanskiy}
\author[1, 2]{\small Artem Klimov}

\affil[1]{\footnotesize Ural Federal University, Ekaterinburg, Russia}
\affil[2]{\footnotesize LLC "Nexus", Ekaterinburg, Russia}

\maketitle
\begin{abstract}
    The transformer architecture has become an integral part of the field of modern neural networks, playing a crucial role in a variety of tasks, such as text generation, machine translation, image and audio processing, among others. There is also an alternative approach to building intelligent systems, proposed by Jeff Hawkins and inspired by the processes occurring in the neocortex. In our article we want to combine some of these ideas and to propose the use of homeostazis mechanisms, such as RFB-kWTA and "Smart"\ Inhibition, in the attention mechanism of the transformer and at the output of the transformer block, as well as conducting an experiment involving the introduction of sparse distributed representations of the transformer at various points. RFB-kWTA utilizes statistics of layer activations across time to adjust the entire layer, enhancing the values of rare activations while reducing those of frequent ones. "Smart"\ Inhibition also uses activation statistics to sample sparsity masks, with rarer activation times are more likely to be activated. Our proposed mechanisms significantly outperform the classical transformer 0.2768 BLEU and a model that only makes use of dropout in the attention mechanism and output of the transformer block 0.3007 BLEU, achieving a score of 0.3062 on the Multi30K dataset.  
\end{abstract}


\section{Introduction}
In 2017, a new architecture was proposed \cite{attention_is_all_you_need}, marking the beginning of a class of architectures named "transformer". The transformer has significantly improved machine translation quality. Subsequently, this model developed into the LLMs known today \cite{mistral, llama, gemma, phi3}. Additionally, researchers have found that models based on the transformer idea can achieve high results in image \cite{vit} and audio \cite{soundtransformer} processing tasks. The high efficiency results of this architecture are due to the self-attention mechanism, which allows to assess the influence of tokens on each other. In other words, consider the sequence in context. Another significant feature is the encoding of token positions in the sequence using harmonic functions, which serves as the primary function of token location. Furthermore, it is noteworthy that both the encoder and decoder utilize the principle of ResNet residual networks \cite{resnet}, making connections not only with the lower layer but also with layers two levels below. This allows for the construction of a network with a high number of layers, thereby avoiding a well-known issue such as a vanishing gradient during training. 

At the same time, the transformer architecture enables parallelization to be carried out much more efficiently compared to well-known recurrent models such as LSTM \cite{lstm} and GRU \cite{gru}. This feature, coupled with significant advances in GPU performance, has made it possible to develop models with billions of parameters and train them on trillions of tokens, which in turn has led to a qualitative breakthrough in language modeling. 
 
Due to its flexibility and universality, the transformer has become an important component in modern ML and is finding increasing application in a variety of tasks \cite{transformeruniversal}. 

On the other hand, in recent years, there has been discussion within the scientific community regarding the Thousand Brains Theory ideas \cite{tbt}. This theory is based on research into the neocortex and the development of biologically likelihood models. These differ significantly from more traditional machine learning approaches, offering an alternate perspective on the creation of artificial intelligence systems. At the same time, a number of ideas can be applied within traditional AI models. In particular, these include:
\begin{enumerate}
	\item Sparse distributed representations.
	\item Self-organization of cells.
\end{enumerate}

Sparse distributed representations can be defined as a tensor with a large number of zero elements. In their papers \cite{sdr, htm, cortical}, the authors demonstrate that the use of SDR leads to:
\begin{enumerate}
	\item Increasing the capacity of the neural network.
	\item Noise resistance.
\end{enumerate}

The use of local and global inhibition in the spatial grouping of hierarchical temporal memory \cite{htm} is inspired by the properties of inhibitory neurons in the neocortex \cite{inhibition1, inhibition2}. The use of inhibition properties allows, on the one hand, to obtain the sparse distributed representations, and on the other hand, to implement the mechanism of unsupervised learning in HTM.  Experimental modeling has shown that the use of these approaches significantly enhances the efficiency of the model.

Previously, inhibition mechanisms were employed in Fukushima's \cite{neocognitron} work on neocognitrons. This model, similar to HTM, exhibits the property of self-organization. 

In this study, we aim to combine the ideas of classical machine learning and the model of the biological neocortex into a unified model based on the classical transformer. Despite the high efficiency of the transformer architecture, we propose an improvement in the self-attention mechanism with the introduction of sparsity by the kWTA method \cite{kwta} with a statistical enhancement of rare activations, which we called Rare Features Boosting kWTA or RFB-kWTA. Subsequently, we demonstrate that integrating this mechanism into the self-attention layer of the transformer allows for improved generalization of the model compared to \cite{attention_is_all_you_need}.

\section{Methods and Materials}

The proposed method is primarily based on two principles: sparsity and self-organization \cite{cortical, neocognitron}.  We achieve sparsity by using kWTA, suppressing the smallest features. 

On the other hand, we hypothesize that the discarded features with the lowest values may contain information that helps the model to generalize patterns in the data, therefore, we propose a homeostatic mechanism that enhances rarer features over time before applying kWTA. This way, the distribution of features is controlled by the model itself, thereby improving its ability to generalize. RFB-kWTA is essentially a simplified representation of the biological self-regulation process of cells in the neocortex. 

\subsection{kWTA}

The essence of kWTA is to discard (zeroing) all values except for a certain proportion of the largest values. Consider a vector $x=[x_1,…,x_n]$ of size $n$. We introduce the sparsity coefficient $s\in(0;1)$ - the proportion of non-zero elements in the vector. Let us assume that a higher sparsity corresponds to a lower value of $s$. The result of kWTA's work is a vector $\widetilde{x} =x\odot m$, where $m=[m_1,…,m_n]$ is a binary vector:

\begin{equation}
    \begin{cases}
        1, x_i \in M \\
        0, otherwise
    \end{cases},
\end{equation}

$M$ is a set consisting of $k=round(s\cdot n)$ of the largest elements of the vector $x$. Next, we will call $m$ the kWTA activation mask. After calculation, the resulting vector $x$ has zero values at the places where $p=n-k$ of the smallest elements are located

\subsection{Application of kWTA in the self-attention layer}

In the paper \cite{attention_is_all_you_need}, one of the main mechanisms is self-attention. The calculation of self-attention can be written as:

\begin{equation}
    SA=Softmax(\frac{QK^T}{D})\cdot V.
\end{equation}

The $SA$ tensor has a size $(B,L,H,D_h)$, where $B$ is the size of the batch, $L$ is the length of the sequence, $H$ is the number of attention heads, $D_h=\frac{D}{H}$ is the size of the representation of each head. 

kWTA is applied to the last measurement of SA:

\begin{equation}
    out=kWTA_s(SA).
\end{equation}

The output is the $out$ tensor, which has the same size as $SA$, while the proportion of non-zero elements in $SA$ is $s$ with an accuracy of rounding error.

\subsection{RFB-kWTA algorithm}
\label{sec:2.3}
The mechanism we propose contains another side: the strengthening of those positions in vector $x$, that were statistically less likely to be activated. In order to implement such a mechanism, we will collect activation statistics at each step of training in a tensor $T$ having size $(H,Q,D_h)$. At each step of the training, activation statistics are calculated by batch for each head, that is, the elements of the tensor $T$ - $t_{h_i,j}=[t_{h_i,j,1},…,t_{h_i,j,D_h}]$:

\begin{equation}
    t_{h_i,j,k}=\sum_{l,p}{M_{l,p,h_i,k}},
\end{equation}

$M$ is the tensor of masks of the same size as $SA$,
$h_i$ is the self-attention head index,
$j$ is the cache position index,
$k$ is the position of the embedding element, 
$l$ is the batch index,
$p$ is the index of the sequence position.

Tensor $T$ is a cache operating on the FIFO (First in First out) principle with size $Q$. At each step of calculating self-attention with kWTA, we use these statistics to calculate the mask $m$ as follows: calculate the total activation statistics for $Q$ steps, that is, the vector $\overline{t}_{h_i}=[\overline{t}_{h_i, 1}, ..., \overline{t}_{h_i, D_h}]$, the elements of which are calculated:

\begin{equation}
    \overline{t}_{h_i}=\sum_{p=1}^{Q}{t_{h_i,p,j}},
\end{equation}

Let the input tensor for the attention block be $X$, with dimensions $(B,L,H,D_h)$. We will calculate the tensor of enhanced activations as follows:

\begin{equation}
    \widetilde{X}_{b,s,h_i,d}=X_{b,s,h_i,d}\odot \frac{max(t_{h_i})-t_{h_i}+min(t_{h_i})}{v},
\end{equation}

where $max(t_{h_i})$ is the maximum element of activation statistics for the $h_i$ head, $min(t_{h_i})$ is the minimum element of activation statistics for the $h_i$ head. Let the statistics values for each head be sorted in descending order, then $v$ is an element with index $k=round(s\cdot n)$.

Note that the calculation of the kWTA remains unchanged, with the result still equal to $\widetilde{x} =x\odot m$, Only the vector for which the mask $m$ is applied changes.

\subsection{"Smart"\ Inhibition}

In addition, we have implemented a method that incorporates stochastic (similar to dropout) and statistical inhibition of activations. The idea is similar to the RFB-kWTA method: we accumulate activation statistics over time, and then sample the mask randomly with probabilities depending on activation statistics.

Similarly to the previous paragraph, we accumulate activation statistics, the size of which is $(H,Q,D_h)$, if applied in self-attention, or $(Q,D_h)$, when applied in the output of the transformer block.

Next, we get a vector (matrix) of probabilities for sampling the mask:

\begin{equation}
    P=\left((a-b)\cdot \frac{\overline{t}-min(\overline{t})}{max(\overline{t})-min(\overline{t})}\right)^{0.83}.
\end{equation}

Here $a=0.99$ is the maximum possible probability, $b=0.01$ is the minimum possible probability. $\overline{t}$ is the aggregated activation statistics, which is calculated as in \ref{sec:2.3}.
Next, to maintain constant thinning, we change the probabilities as follows:

\begin{equation}
    P=\begin{cases}
        P+s-median(P), |median(P)-s| > \delta \\
        P, otherwise
    \end{cases},
\end{equation}

Next, we use tiling (tile function in PyTorch \cite{pytorch} or Numpy \cite{numpy}), adjusting the size of $P$ to $X$. After that, using the Bernoulli distribution, a binary mask is formed, of the same size as $X$. The output of the layer is $X\odot M$, $M$ is the sampled mask.

\subsection{Memorization speed estimation}

In the section with the results, we will pay attention to the speed of memorization of training data by the model. To numerically estimate this value, we introduce the characteristic:

\begin{equation}
    IMI=\frac{\sum_{i=1}^{E-1}{f_{i+1}-f_i}}{2(E-1)}.
\end{equation}

$f_i$ is the metric value for the $i$-th epoch, $E$ is the number of training epochs.
This value is the ratio of the area under the curve of the metric trained by the model to the area under the curve of the ideal model - a model that remembers the entire training sample for the first epoch.

\subsection{Method of conducting the experiment}

To test the effectiveness of the proposed algorithm, we used the Multi30k en-de dataset \cite{multi30k}. The tokenizers was taken from the dataset itself. The training set contains 29k pairs of sentences, and the test and validation sets each contain 1,024 pairs.

We trained the models for 181k steps on the RTX 4090 and 3080Ti, the parameters of the models and training will be given later in the results section.

To train the model, we used Adam optimizer \cite{adam} with $lr=0.0001$ and Cross-Entropy as a loss function.

To evaluate the result of the model, the BLEU metric was chosen \cite{bleu}, which was calculated on the lines generated by the token-by-token transformer. To get the best results, 5 checkpoints were saved every 30 epochs: the best loss and BLEU values for train and validation, as well as the checkpoint from the last epoch.

\section{Results}

\begin{figure}[t]
	\centering
        \captionsetup{justification=centering}
	\includegraphics[scale=0.5]{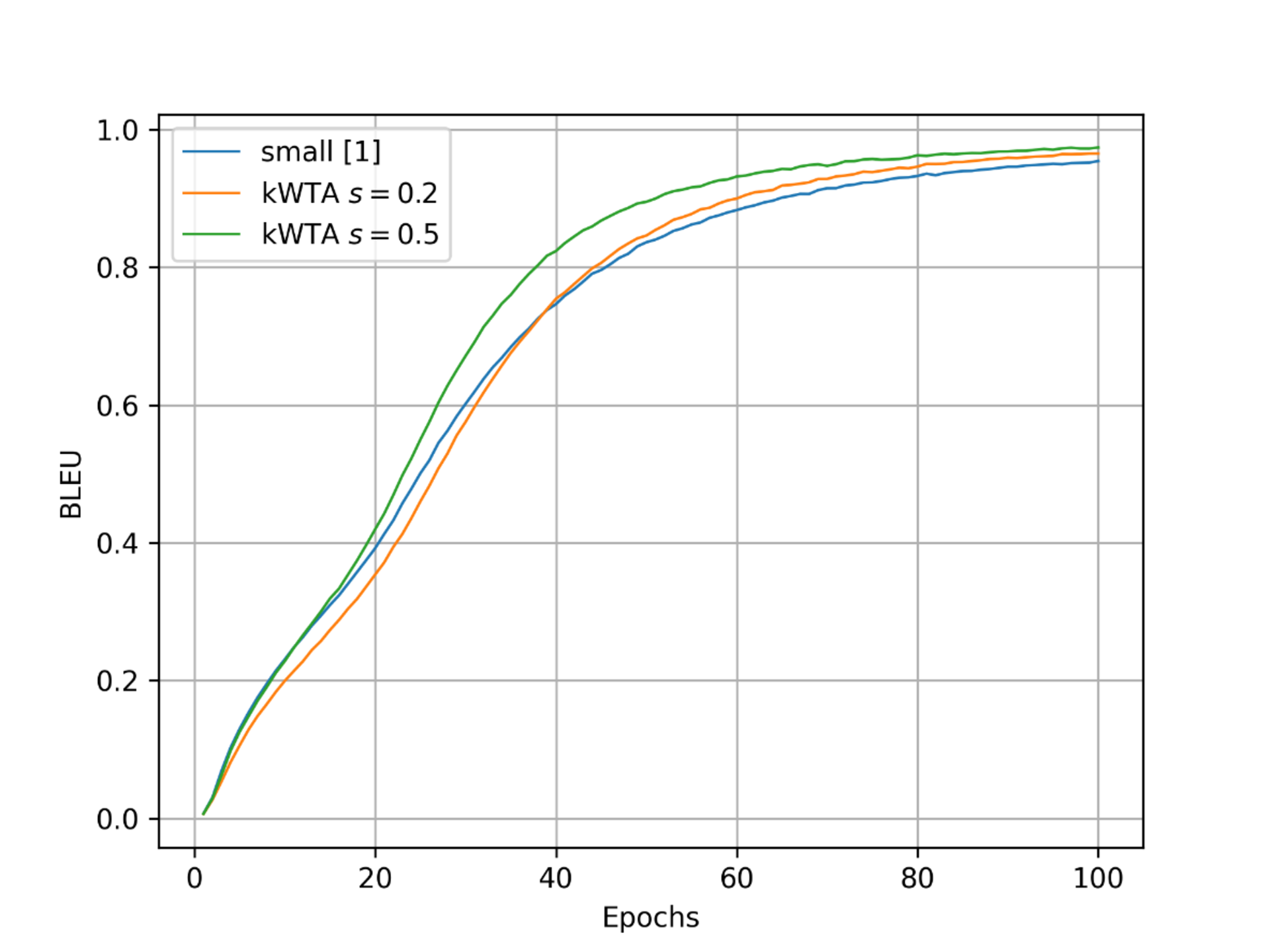}
	\caption{BLEU graph based on training data for small, small-kWTA $s=0.2$ and small-kWTA $s=0.5$ models.}
	\label{fig:fig1}
\end{figure}

In this section, we will first demonstrate the effect of kWTA compared to a classic transformer. We will analyze the memorization and generalization abilities of both models and provide training schedules, on which we will compare the differences between the models. 

Then we investigate the behavior of the RFB-kWTA and "Smart"\ Dropout algorithms on the same data. We will compare their results with those of models based on Attention Is All You Need, in order to demonstrate the benefits of our proposed methods.

\subsection{Parameters of the tested models }

In this section, we will present three types of tested models, which for simplicity are called small (23.1M), base (68.3M) and big (224.6M). All parameters are given in Table \ref{tab:table1}. The models given are models reproduced from work \cite{attention_is_all_you_need} with the same parameters, except for the small model, which was not in the original work.

\begin{table}
	\caption{Model parameters}
	\centering
	\begin{tabular}{ccccccc}
		\toprule
		Name     & Params, $\times10^6$     & $D$ & $H$ & $D_{ff}$ & $N_{blocks}$ & Dropout rate \\
		\midrule
		Small     & 23.1    & 256 & 4 & 1024 & 6 & 0.1 \\
		  Base     & 68.3   & 512 & 8 & 2048 & 6 & 0.1 \\
		  Big     & 224.6   & 1024 & 16 & 4096 & 6 & 0.1 \\
		\bottomrule
	\end{tabular}
	\label{tab:table1}
\end{table}

During the experiments, we implemented the proposed mechanisms at two points of the model: in the self-attention and after the residual connection at the output of the encoder and decoder blocks. In addition to the proposed mechanisms, we also used dropout at these points.

\subsection{The influence of kWTA on memorization and generalization}

We assume that the use of exclusively kWTA on the one hand leads to an improvement in memorization by the model, and on the other - to a deterioration in generalization. The improvement of memorization occurs due to the selection of the strongest features while simultaneously removing the weak ones. The deterioration of generalization occurs due to the loss of some of the features. 

\begin{figure}[h!]
	\centering
        \captionsetup{justification=centering}
	\includegraphics[scale=0.5]{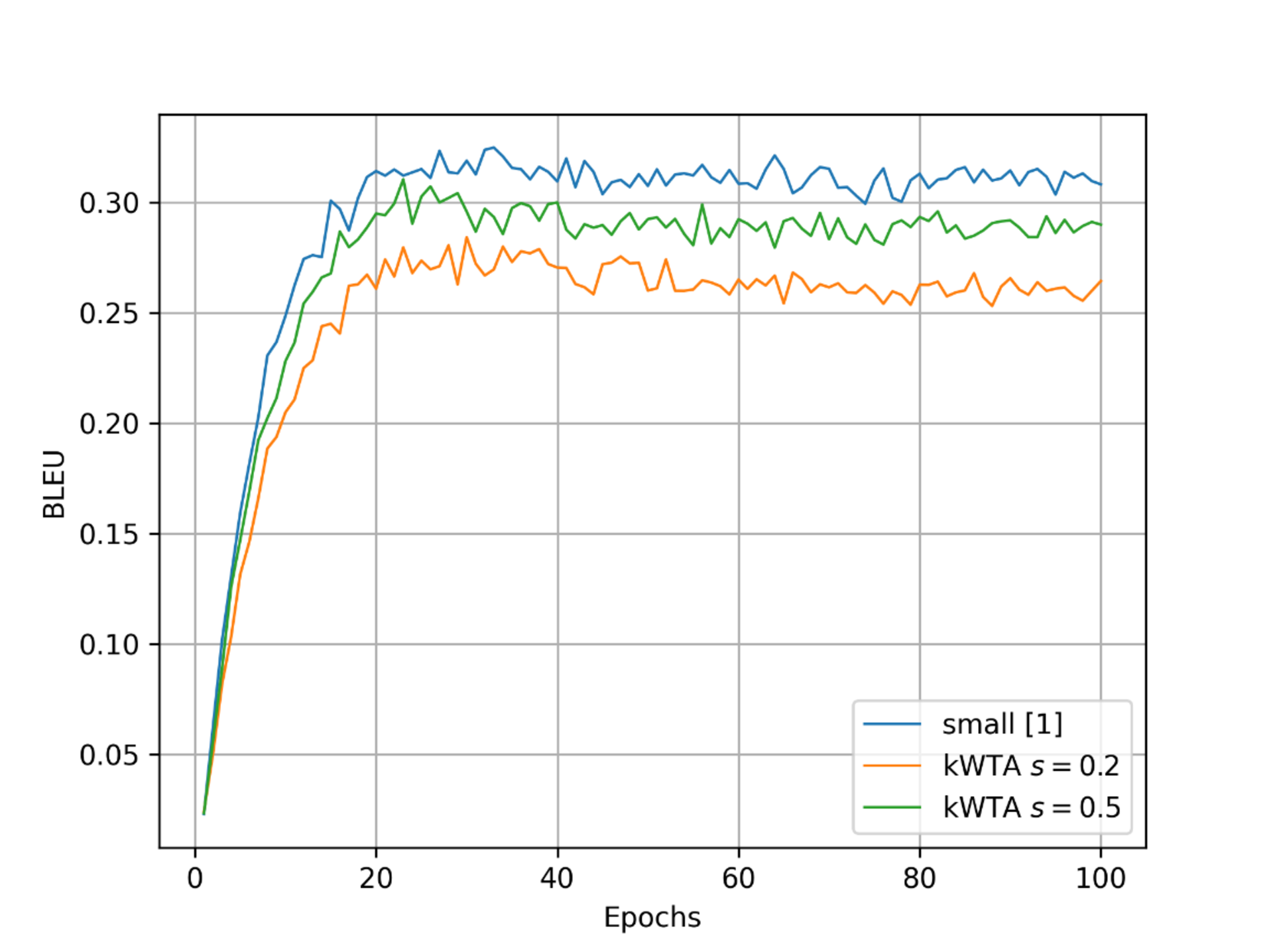}
	\caption{BLEU graph based on validation data for models small, small with kWTA $s=0.2$ and small with kWTA $s=0.5$}
	\label{fig:fig2}
\end{figure}

To confirm the hypothesis of improved memorization by a model using kWTA in the self-attention layer without using time-rare feature boosting, we conducted an experiment in which we trained a model from \cite{attention_is_all_you_need} and a model with kWTA in the self-attention layer with different coefficients $s$.

The training was conducted over 45k steps. The graph of the BLEU metric for the small model in training and validation samples is shown in Figure \ref{fig:fig1} and Figure \ref{fig:fig2}.

Figure \ref{fig:fig1} shows that the curve for the kWTA model with a coefficient $s=0.5$ is higher than the model from \cite{attention_is_all_you_need} throughout the training. The results are shown in Table \ref{tab:table2}. 

\begin{table}[h]
	\caption{Best model results on train and validation}
	\centering
	\begin{tabular}{ccccccc}
		\toprule
		Model     & $BLEU_{train}$  & $BLEU_{val}$ & $IMI_{train}$ \\
		\midrule
		Small     & 0.954   & 0.325 & 0.705  \\
		  Small kWTA $s=0.2$     & 0.965   & 0.284 & 0.728  \\
		  Small kWTA $s=0.5$     & 0.974   & 0.310 & 0.743  \\
		\bottomrule
	\end{tabular}
	\label{tab:table2}
\end{table}

From all the above results, we can conclude that when using kWTA in the self-attention layer, there is an increase in the ability of the model to memorize training data and a deterioration in generalization. 

\subsection{RFB-kWTA Testing}

\begin{figure}[]
	\centering
        \captionsetup{justification=centering}
	\includegraphics[scale=0.7]{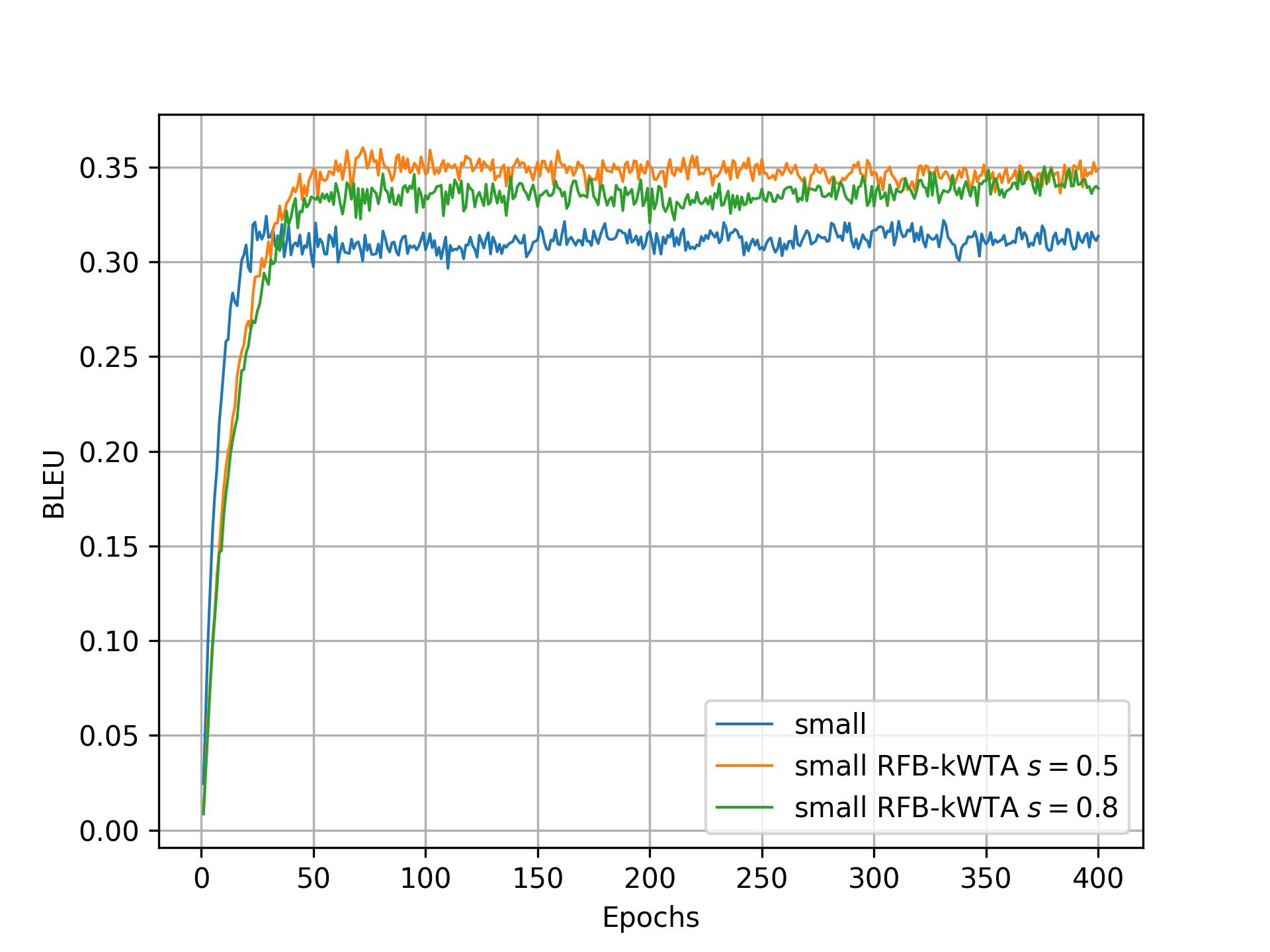}
	\caption{BLEU graph based on validation data for models small, small RFB-kWTA $s=0.5$ and small RFB-kWTA $s=0.8$}
	\label{fig:fig3}
\end{figure}

In this experiment, we implemented RFB- kWTA in self-attention and dropout in the interblock connection.  We will compare small models with each other: small, small RFB-kWTA with a coefficient $s=0.5$ and small RFB- kWTA with a coefficient $s=0.8$. The BLEU graph on validation is shown in Figure  \ref{fig:fig3}.

It can be seen from Figure \ref{fig:fig3} that the introduction of sparsity by the RFB-kWTA method can significantly improve the generalization of the model.

\begin{figure}[]
	\centering
        \captionsetup{justification=centering}
	\includegraphics[scale=0.7]{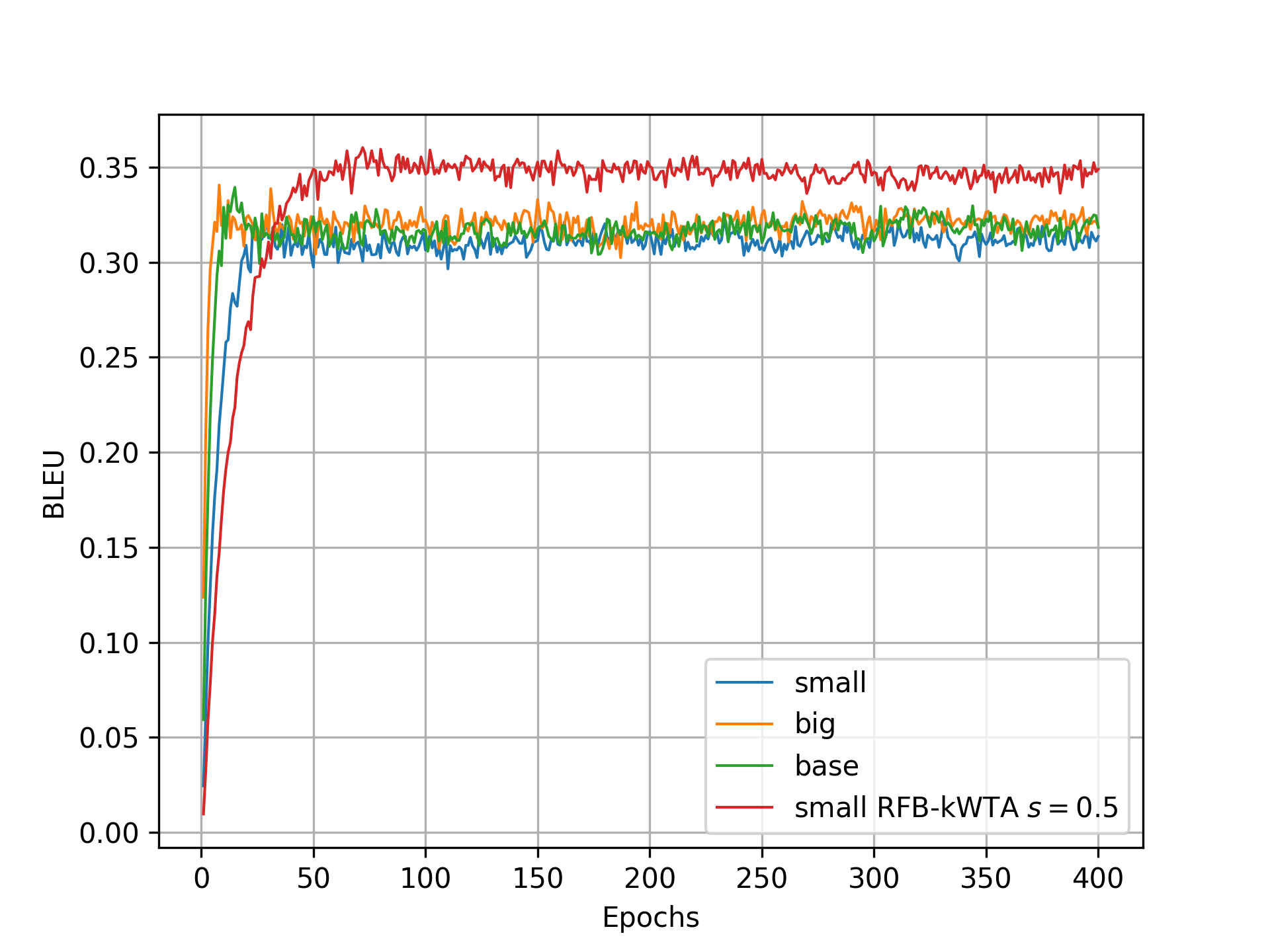}
	\caption{BLEU graph based on validation data for models small, base, big and small RFB-kWTA $s=0.5$}
	\label{fig:fig4}
\end{figure}

We also present a comparison with models with a larger number of parameters, that is, with the base and big models. This comparison is shown in Figure \ref{fig:fig4}.

Figure \ref{fig:fig4} shows that the big model, containing almost 10 times more trainable parameters, has a lower generalization quality compared to the small RFB-kWTA $s=0.5$.

\subsection{Results of other tests}

In this section, we present a summary table of all the results obtained during the study. The following experiments were conducted: 

\begin{enumerate}
	\item Models from \cite{attention_is_all_you_need}.
	\item RFB-kWTA in self-attention and Dropout in the interblock connection.
	\item Dropout in self-attention and Dropout in the interblock connection. 
	\item RFB-kWTA in self-attention and "Smart"\ Inhibition in the interblock connection. 
	\item "Smart"\ Inhibition in self-attention and "Smart"\ Inhibition in the interblock connection.
\end{enumerate}

Due to the large number of possible experiments, we were able to consider a limited number of combinations, so we present the results of all the tests in a single summary table.

The BLEU score based on test data was built in token-by-token mode. All experimental results are shown in Table \ref{tab:table3}. The best test results are shown in color.

\small
\begin{table}
	\caption{Test results for all models. (A) - models from Attention is All You Need, (B) - models with RFB-kWTA in the attention mechanism and Dropout at the output of the transformer block, (C) - model with Dropout in the attention mechanism and at the output of the transformer block, (D) - models with "Smart"\ Inhibition in the mechanism attention and at the output of the transformer block, (E) - models with RFB-kWTA in the attention mechanism and "Smart"\ Inhibition at the output of the transformer block.}
	\centering
	\begin{tabular}{C{2cm}C{2cm}C{2cm}C{2cm}C{2cm}C{2cm}}
		\toprule
	        & Model & $Q_{Att}$ & $Q_{BO}$ & $s$ & $BLEU$ \\
            \midrule
		\multirow{3}{*}{(A)} & Small & - & - & - & 0.2727   \\
		  & Base & - & - & - & 0.2768  \\
		  & Big  & - & - & - & 0.2751  \\
            \midrule
            \multirow{12}{*}{(B)} & Small & 256 & - & 0.5 & 0.2926   \\
		  & Small & 256 & - & 0.7 & 0.2898  \\
		  & \cellcolor{green!15}Small & \cellcolor{green!15}256 & \cellcolor{green!15}- & \cellcolor{green!15}0.8 & \cellcolor{green!15}0.3025  \\
		  & Small & 256 & - & 0.9 & 0.2967  \\
		  & Base & 256 & - & 0.5 & 0.2848  \\
		  & Base & 256 & - & 0.7 & 0.2918  \\
		  & Base & 256 & - & 0.8 & 0.2888  \\
		  & \cellcolor{green!30}Base & \cellcolor{green!30}256 & \cellcolor{green!30}- & \cellcolor{green!30}0.9 & \cellcolor{green!30}0.3046  \\
		  & Big & 256 & - & 0.5 & 0.2863  \\
		  & Big & 256 & - & 0.7 & 0.2858  \\
		  & Big & 256 & - & 0.8 & 0.2913  \\
		  & Big & 256 & - & 0.9 & 0.2927  \\
            \midrule
            (C) & Base & - & - & - & 0.3007 \\
            \midrule
            \multirow{4}{*}{(D)} & Base & 256 & 1 & 0.9 & 0.2942 \\
            & \cellcolor{green!45}Base & \cellcolor{green!45}256 & \cellcolor{green!45}16 & \cellcolor{green!45}0.9 & \cellcolor{green!45}0.3062 \\
            & Base & 256 & 256 & 0.9 & 0.2948 \\
            & Base & 256 & 1024 & 0.9 & 0.2854 \\
            \midrule
            \multirow{4}{*}{(E)} & Big & 256 & 1 & 0.9 & 0.2891 \\
            & Big & 256 & 4 & 0.9 & 0.2913 \\
            & Big & 256 & 16 & 0.9 & 0.2941 \\
            & Big & 256 & 512 & 0.9 & 0.2868 \\
		\bottomrule
	\end{tabular}
	\label{tab:table3}
\end{table}

\section{Discussion}

During the study, we saw two important effects:

\begin{enumerate}
	\item Thinning the representation in self-attention using kWTA improves the model's memorization of training data, but worsens generalization. The memorization effect is observed due to the selection of the most "bright" features, allowing the model to memorize the training sample faster. At the same time, the extraction of less "bright" signs leads to the loss of information, which contains "subtle" patterns leading to generalization. 
	\item The RFB mechanism in combination with Dropout has a positive effect on the generalizing ability of the model. In order to allow "subtle" patterns to periodically appear among the signs, we introduced a mechanism of homeostazis based on activation statistics. This mechanism makes it possible to enhance signs that are rare in time, which may contain those very "subtle" patterns. However, substituting dropout with “smart” inhibition allows for further improvement in the quality of the model. 
\end{enumerate}

The mechanisms proposed by us, due to their self-regulation, outperform the classical transformer in the task of machine translation. At the same time, all models incorporating a homeostazis mechanism, even those with $s=0.5$ have proven to be superior to the classic transformer model.

Due to limited computing resources and a large number of experiments, we have not established a dependence of quality on the size of the $Q_{Att}$. In the future, we plan to conduct such a study. 

The results obtained can be useful for constructing models with attention blocks and classical models consisting of an encoder and a decoder, however, there is a need to verify the applicability of the mechanism for other tasks (for example, image processing). We assume that the method is generalizable to any other tasks.

The proposed model showed a significantly higher result than the classic transformer, however, we did not see the expected victory of the large model over the smaller ones. We attribute this to the redundancy of the large model for this task, since the dataset contains a total of 30k samples. Nevertheless, the base model performs better than the small one, and this allows us to conclude that the method is applicable in architectures with a large number of parameters.

\section{Conclusion}

In this study, we proposed a self-regulation method that allows the model to significantly improve the quality of generalization. In particular, we have shown that the small model, which has about 10 times fewer trainable parameters compared to the big model of the classical transformer, significantly outperforms it as a generalization - 0.3025 BLEU versus 0.2751 BLEU. We believe that such an increase was obtained due to the self-regulating strengthening of insignificant signs in self-attention, which are nevertheless important for the formation of rules by the model. At the same time, we got the maximum result from the base model with a combination of "Smart"\ Inhibition in both cases. 

We have also identified an important correlation between the sparsity of the kWTA and the quality of memorization and generalization. At a certain level of sparsity, the quality of memorization is achieved better than in the classical transformer with small losses of generalization. Next, we intend to test the proposed mechanism in other tasks and models of architectures. Additionally, an interesting experiment may be to combine multiple attention heads with a single activation statistic.

\bibliographystyle{unsrtnat}
\bibliography{references}

\end{document}